\documentclass[letterpaper]{article} 
\usepackage{aaai25}  
\usepackage{times}  
\usepackage{helvet}  
\usepackage{courier}  
\usepackage[hyphens]{url}  
\usepackage{graphicx} 
\usepackage{natbib}  
\usepackage{caption} 
\usepackage{algorithm}

\usepackage{newfloat}
\usepackage{listings}
\DeclareCaptionStyle{ruled}{labelfont=normalfont,labelsep=colon,strut=off} 
\floatstyle{ruled}
\newfloat{listing}{tb}{lst}{}
\floatname{listing}{Listing}
\title{Towards Efficient Risk-Sensitive Policy Gradient:
An Iteration Complexity Analysis}

\author {
    Rui Liu,
    Anish Gupta,
    Erfaun Noorani,
    Pratap Tokekar
}
\affiliations {
    University of Maryland, College Park \\
\{ruiliu, agupta96, enoorani, tokekar\}@umd.edu
}

\usepackage{multicol}
\usepackage{graphics} 
\usepackage{amsmath} 
\usepackage{amssymb}  
\usepackage{algorithm}
\usepackage{multirow}
\usepackage{algpseudocode}
\usepackage{mathtools}
\usepackage{subcaption}
\usepackage[dvipsnames]{xcolor}

\usepackage{epsfig} 
\usepackage{times} 
\usepackage{amsmath} 
\usepackage{amssymb}  
\usepackage{algpseudocode}

\newcommand{\cO}{\mathcal{O}}
\newcommand{\dotprod}[1]{\left< #1\right>} 
\newcommand{\norm}[1]{ \left\| #1 \right\|}      
\newcommand{\eqdef}{=\vcentcolon}
\newcommand{\defeq}{\stackrel{\text{def}}{=}}
\newcommand{\hnabla}{\widehat{\nabla}}
\newcommand{\E}[1]{\mathbb{E}\left[#1\right] } 

\newcommand{\EE}[2]{\mathbb{E}_{#1}\left[#2\right] } 
\usepackage{tikz}
\newcommand*\circled[1]{\tikz[baseline=(char.base)]{
            \node[shape=circle,draw,inner sep=.5pt] (char) {#1};}}
            
\newtheorem{assumption}{Assumption}
\newtheorem{theorem}{Theorem}
\newtheorem{lemma}{Lemma}

\newtheorem{coro}{Corollary}

\def\cA{{\mathcal{A}}}

\def\cO{{\mathcal{O}}}
\def\cP{{\mathcal{P}}}

\def\cR{{\mathcal{R}}}
\def\cS{{\mathcal{S}}}

\def\sR{{\mathbb{R}}}

\begin{document}

\maketitle

\begin{abstract}
Reinforcement Learning (RL) has shown exceptional performance across various applications, enabling autonomous agents to learn optimal policies through interaction with their environments. However, traditional RL frameworks often face challenges in terms of iteration efficiency and safety. Risk-sensitive policy gradient methods, which incorporate both expected return and risk measures, have been explored for their ability to yield safe policies, yet their iteration complexity remains largely underexplored. In this work, we conduct a rigorous iteration complexity analysis for the risk-sensitive policy gradient method, focusing on the REINFORCE algorithm with an exponential utility function. We establish an iteration complexity of $\cO(\epsilon^{-2})$ to reach an $\epsilon$-approximate first-order stationary point (FOSP). Furthermore, we investigate whether risk-sensitive algorithms can achieve better iteration complexity compared to their risk-neutral counterparts. Our analysis indicates that risk-sensitive REINFORCE can potentially converge faster. To validate our analysis, we empirically evaluate the learning performance and convergence efficiency of the risk-neutral and risk-sensitive REINFORCE algorithms in multiple environments: CartPole, MiniGrid, and Robot Navigation. Empirical results confirm that risk-sensitive cases can converge and stabilize faster compared to their risk-neutral counterparts. More details can be found on our website\footnote{\url{https://anonymous.4open.science/w/riskrl/}}.
\end{abstract}

 
\section{Introduction}
Reinforcement Learning (RL) is the problem of learning optimal policies through interactions with an environment \cite{sutton1999policy,kaelbling1996reinforcement}. RL has shown remarkable success in a wide range of applications, e.g., board and video game playing \cite{silver2016mastering, mnih2013playing}, and LLM reasoning \cite{zheng2025learning}. However, it is widely acknowledged that classical RL is both lacking in safety and falls short in terms of iteration efficiency \cite{casper2023open, almahamid2021reinforcement}. One reason is that standard RL only takes expected return into consideration. Risk-sensitive RL algorithms \cite{mihatsch2002risk, shen2014risk, berkenkamp2017safe} mitigate these issues by taking into account not only the expected value of performance but also its variability. This allows for adjusting the balance between the expected return and variability. Consideration of risk and safety is crucial in high-stake and safety-critical applications, such as finance \cite{filos2019reinforcement, charpentier2021reinforcement}, autonomous driving \cite{zhang2021safe, liu2025aukt, liu2025caml} and robotics \cite{majumdar2017risk, liu2024adaptive, liu2024imrl}. In these practical scenarios,  it is not enough to merely optimize for expected returns. Various risk measures, such as Conditional Value-at-Risk (CVaR) \cite{qiu2021rmix, prashanth2022risk,zhou2022risk}, Optimized Certainty Equivalents (OCE) \cite{lee2020learning}, chance constraints~\cite{shek2025localize}, and exponential utility function \cite{mihatsch2002risk, fei2020risk, eriksson2019epistemic, liu2023data, prashanth2022risk, noorani2021risk, noorani2022risk}, have been used to incorporate safety into RL algorithms. The safety of policies derived from risk-sensitive RL algorithms employing an exponential utility function has been both analytically established and empirically validated, as demonstrated by Noorani et al. \cite{noorani2022risk}.
 
While prior works have developed risk-sensitive RL algorithms based on these risk measures, their iteration complexity, defined as the number of iterations required to reach a satisfactory solution, has received limited attention. Understanding iteration complexity \cite{kakade2003sample, dann2015sample, lattimore2013sample} is essential, as it provides insights into the efficiency of risk-sensitive RL and informs the development of more efficient algorithms. In practical scenarios, if a risk-sensitive algorithm can converge faster or exhibit better stability while managing safety effectively, it can offer significant advantages over traditional methods. For instance, a risk-sensitive policy that avoids overly risky actions may reduce the likelihood of catastrophic failures, thereby minimizing the number of iterations needed for recovery. This translates to more efficient training and lower computational costs. Bejarano et al. \citet{bejarano2024safety} explored a similar idea in a recent study, where an augmented safety filter in RL training, rather than risk sensitivity, prevents the exploration of unsafe states, leading to improved performance and better sample efficiency. Motivated by these considerations, we investigate the iteration complexity of risk-sensitive RL algorithms, and further seek to address the key question: \emph{Can risk-sensitive algorithms potentially exhibit improved iteration complexity compared to standard methods?}



In this paper, we focus on the Policy Gradient (PG) method REINFORCE \cite{williams1992simple,sutton1999policy,baxter2001infinite} and its risk-sensitive variant \cite{noorani2021risk, noorani2022risk}, which incorporates an exponential utility function into the learning objective. Although previous studies \cite{papini2018stochastic, xu2020improved, xu2019sample, papini2021safe, yuan2022general} have analyzed the iteration complexity of the standard risk-neutral REINFORCE algorithm, few have explored the iteration complexity of the risk-sensitive counterpart. For instance, \citet{yuan2022general} established convergence rates and iteration complexity bounds for the standard risk-neutral REINFORCE algorithm. However, they did not examine the risk-sensitive algorithm.


\begin{table}[ht]
\centering
\small
\setlength{\tabcolsep}{1.5mm} 
\begin{tabular}{lccl}
\hline
Reference & Type & Guarantee & Bound \\ \hline
\citet{papini2018stochastic} & RN & FOSP & $\cO(\epsilon^{-2})$  \\
\citet{xu2020improved} & RN & FOSP & $\cO(\epsilon^{-\frac{5}{3}})$ \\ 
\citet{xu2019sample} & RN & FOSP & $\cO(\epsilon^{-\frac{3}{2}})$ \\
\citet{papini2021safe} & RN & FOSP & $\cO(\epsilon^{-2})$ \\
\citet{yuan2022general} & RN & FOSP & $\cO(\epsilon^{-2})$ \\
Ours & RS & FOSP &  $\cO(\epsilon^{-2})$ \\ \hline
\end{tabular}%
\caption{\textbf{Iteration complexities of previous standard REINFORCE and our risk-sensitive REINFORCE}. RN denotes risk-neutral and RS denotes risk-sensitive.}
\label{tab:bound}
\end{table}



Unlike prior works, we aim to analyze the iteration complexity of the risk-sensitive REINFORCE algorithm. Specifically, we establish an iteration complexity bound for reaching an $\epsilon$-approximate first-order stationary point (FOSP), ensuring that $\E{\norm{\nabla J_\beta(\theta)}} \leq \epsilon$, where $J_\beta$ denotes the risk-sensitive objective function parameterized by $\theta$ with a risk-sensitive parameter $\beta$. We present in Table \ref{tab:bound} iteration complexity results from previous studies on standard risk-neutral REINFORCE and our findings for risk-sensitive REINFORCE. 

Beyond characterizing the iteration complexity of risk-sensitive REINFORCE, we compare it against its risk-neutral counterpart as the baseline. We begin by empirically analyzing the learning performance and convergence efficiency of both risk-neutral and risk-sensitive REINFORCE algorithms in the CartPole \cite{1606.01540} environment. Our analysis indicates that, under appropriate risk-sensitive parameters, the risk-sensitive algorithm can achieve faster convergence than the standard risk-neutral REINFORCE. To validate our analysis, we further conduct experiments in a holonomic robot navigation environment and a MiniGrid \cite{MinigridMiniworld23} environment. Empirical results confirm that risk-sensitive policies can not only converge faster but also exhibit more stable learning behavior compared to their risk-neutral counterparts, aligning with our analysis.

Overall, we contribute in the following aspects and we summarize these contributions below:
\begin{itemize}
    \item We conduct a comprehensive analysis of the iteration complexity for the risk-sensitive REINFORCE algorithm, which exhibits an iteration complexity of $\cO(\epsilon^{-2})$. 
    \item We formally compare the iteration complexity of risk-sensitive and risk-neutral REINFORCE algorithms. Our analysis demonstrates that, for appropriate risk-sensitive parameters $\beta$, the risk-sensitive algorithm can potentially converge faster than its risk-neutral counterpart. We establish the existence of such $\beta$ values. 
    \item We empirically validate the analysis through various experiments, including CartPole, Minigrid, and holonomic robot navigation. The empirical results confirm that risk-sensitive algorithms can converge faster and achieve greater stability. 
\end{itemize}


\section{Related Work}
\subsection{Safe Reinforcement Learning}
Safe RL has focused on integrating safety or risk measures into the decision-making process to develop policies that are safe to variability in returns. Qiu et al. \cite{qiu2021rmix} introduced a method incorporating the CVaR measure to mitigate reward randomness and environmental uncertainty in multi-agent RL (MARL). Lee and Singh \cite{lee2020learning} investigated the generalization properties of risk-sensitive RL, where the objective is formulated using optimized certainty equivalents (OCE). Fei et al. \cite{fei2020risk} studied risk-sensitive RL in episodic Markov decision processes with unknown transition dynamics, aiming to optimize the total rewards under the exponential utility risk measure. Eriksson and Dimitrakakis \cite{eriksson2019epistemic} proposed a framework for RL in uncertain environments by leveraging preferences encoded via the exponential utility function, where risk aversion or seeking behavior can be adjusted through a risk-sensitive parameter. Noorani and Baras \cite{noorani2021risk} introduced risk-sensitive REINFORCE, a Monte Carlo policy gradient algorithm based on an exponential performance criterion.

Despite these advances, most prior works in safe RL have primarily focused on algorithm design and empirical performance, with limited attention to computational efficiency. In particular, the iteration complexity of safe RL algorithms is less studied. Our work addresses this gap by extending the iteration complexity analysis to the risk-sensitive REINFORCE algorithm, providing insights into how risk-sensitive parameters influence convergence rates and demonstrating that risk-sensitive methods can achieve improved efficiency compared to their risk-neutral counterparts.

\subsection{Iteration Complexity and Convergence Analysis}
Iteration complexity analysis plays a crucial role in understanding the efficiency of RL algorithms, particularly in policy optimization methods. For risk-neutral policy gradient methods, several studies have established convergence guarantees and iteration complexity bounds. Papini et al. \cite{papini2018stochastic} introduced the stochastic variance-reduced policy gradient (SVRPG) method, which requires $\cO(\epsilon^{-2})$ iterations to achieve $\norm{\nabla J(\theta)} \leq \epsilon$. Xu et al. \cite{xu2020improved} provided an improved convergence analysis of SVRPG and show an iteration complexity of $\cO(\epsilon^{-\frac{5}{3}})$ to achieve an $\epsilon$-approximate first-order stationary point (FOSP). Subsequently, Xu et al. \cite{xu2019sample} proposed a SRVRPG algorithm that improves this iteration complexity to $\cO(\epsilon^{-\frac{3}{2}})$. Papini et al. \cite{papini2021safe} proved the $\cO(\epsilon^{-2})$ iteration complexity for REINFORCE. Yuan et al. \cite{yuan2022general} achieved an $\cO(\epsilon^{-2})$ iteration complexity for the exact gradient of the REINFORCE algorithm aimed at reaching a FOSP.

However, the iteration complexity of risk-sensitive policy gradient methods remains largely unexplored. While empirical studies suggest that incorporating risk measures (e.g., via the exponential utility function as in \cite{noorani2021risk, noorani2022risk}) may lead to more stable and potentially faster convergence, a rigorous analysis has been missing. In this work, we bridge this gap by conducting a detailed iteration complexity analysis, focusing on the risk-sensitive REINFORCE algorithm. Additionally, we provide a comparative analysis between risk-neutral and risk-sensitive REINFORCE, identifying the conditions under which risk-sensitive REINFORCE can potentially achieve convergence with fewer iterations than its risk-neutral counterpart.

\section{Preliminaries}
\subsection{Markov Decision Process (MDP)} \label{sec: mdp}
We examine a Markov Decision Process (MDP) characterized by the tuple $\{\cS, \cA, \cP, r, \gamma, \rho\}$. In this setup, $\cS$ represents the state space, $\cA$ is the action space, and $\cP$ is the transition model. The transition model, $\cP(s' \mid s, a)$, denotes the probability of transitioning from state $s$ to $s'$ when taking action $a$. The reward function, denoted as $r(s, a)$, produces bounded rewards in the range of $[r_{\min}, r_{\max}]$ for state-action pairs $(s, a)$. The parameters $\gamma \in [0, 1)$ and $\rho$ represent the discount factor and the initial state distribution, respectively. The agent's behavior is captured through a policy $\pi$, which resides in the space of probability distributions over actions at each state. This is represented as $\pi(a\mid s)$. We define the probability density $p(\tau \mid \pi)$ for a single trajectory $\tau$ being generated under policy $\pi$ as follows: 
\begin{equation}
p(\tau\mid\pi) = \rho(s_0)\prod_{t=0}^\infty\pi(a_t\mid s_t)\cP(s_{t+1}\mid s_t,a_t).
\end{equation}

Let $\cR(\tau) \eqdef \sum_{t=0}^\infty\gamma^t r(s_t, a_t)$ be the total discounted rewards accumulated along trajectory $\tau$. We define the risk-neutral expected return of $\pi$ as: 
\begin{equation} 
J(\pi) \defeq \EE{\tau \sim p(\cdot\mid\pi)}{\cR(\tau)}.
\end{equation}

\subsection{Risk-Neutral REINFORCE Algorithm} \label{sec:reinforce}
Policy gradient (PG) methods employ gradient ascent in the parameter space to optimize policies that maximize the expected return. In this work, we focus on the REINFORCE algorithm as the chosen PG method. We consider stochastic and parameterized policies, denoted as $\pi_{\theta}$ with parameter $\theta$. Let's define $J(\pi_{\theta})$ as $J(\theta)$, and we express the probability of a trajectory given $\theta$ as $p(\tau\mid\pi_{\theta}) = p(\tau\mid\theta)$. 

We denote the corresponding optimal expected return as $J^* \defeq J(\theta^*)$. We express the gradient $\nabla J(\theta)$ of the expected return as follows: 
\begin{align} \label{eq:GD}
\nabla J(\theta) &= \int\cR(\tau)\nabla p(\tau\mid\theta)d\tau \\
& = \EE{\tau}{\sum_{t=0}^\infty\nabla_{\theta}\log\pi_{\theta}(a_{t} \mid s_{t}) \sum_{t'=0}^\infty\gamma^{t'} r(s_{t'}, a_{t'})} \\
&= \EE{\tau}{\sum_{t=0}^{\infty} \nabla_\theta \log \pi_\theta(a_t|s_t)R(t)},
\end{align}
where we define $\displaystyle R(t) \defeq \sum_{t'=t}^\infty\gamma^{t'} r(s_{t'}, a_{t'})$ as the discounted rewards-to-go.


Following the preliminaries of the MDP and the risk-neutral REINFORCE algorithm, we introduce the following common assumption, which establishes the smoothness properties of the value function and policy parameterization.

\begin{assumption}[Lipschitz Smoothness] \label{ass: smooth}
    There exists $L \in \mathbb{R}^+$ such that, for all $\theta, \theta' \in \sR^d$ with $l_2$ norm, we have:
    \begin{equation} \label{eq:smooth}
        \norm{\nabla J(\theta) - \nabla J(\theta')} \leq L \norm{\theta-\theta'},
    \end{equation}
\end{assumption}
where the function $J(\theta)$ is the expected return parameterized by $\theta$.

\section{Risk-Sensitive REINFORCE Algorithm}
After discussing the risk-neutral REINFORCE algorithm, we study the risk-sensitive REINFORCE. In this work, we adopt the exponential utility function to incorporate risk into the performance objective, leveraging its computational efficiency and mathematical tractability. We transition from the risk-neutral objective $J$ to the risk-sensitive objective $J_\beta$, where $\beta \neq 0$ is the risk-sensitivity parameter. The risk-sensitive algorithm seeks to maximize the following objective:
\begin{eqnarray}
J_\beta(\pi) \eqdef \EE{\tau \sim p(\cdot \mid \pi)}{\frac{1}{\beta} e^{\beta \cR(\tau)}}, 
\end{eqnarray}

where $\cR(\tau)$ denotes the cumulative discounted rewards along a trajectory $\tau$ as in Section \ref{sec: mdp}.

Some prior works \cite{fei2020risk, hau2023entropic} have employed an objective of the entropic risk measure, defined as $\frac{1}{\beta}\log\big(\EE{\tau \sim p(\cdot \mid \pi)}{e^{\beta \cR(\tau)}}\big)$. From an optimization perspective, this objective yields the same optimal policy as our objective $J_\beta(\pi)$, since both share the same \textit{argmax} due to the strictly increasing nature of the logarithm function. Consequently, the optimal policy is given by $\pi^* = \arg\max_\pi J_\beta (\pi)$. 

The risk-sensitive objective optimizes not only the expected return but also the variance that influences decision-making under uncertainty \cite{noorani2021risk}. In cases where $\beta$ is negative (reflecting risk-averse behavior), maximizing $J_\beta(\pi)$ is equivalent to simultaneously maximizing the expected return and minimizing the variance of the return, which helps stabilize learning. Conversely, when $\beta$ is positive (indicating risk-seeking behavior), maximizing $J_\beta(\pi)$ becomes equivalent to maximizing both the expected return and the variance of the return.

We express the gradient of the objective for the risk-sensitive REINFORCE as follows:

\begin{align} \label{eq:risk-pg}
\vspace{-10pt}
\nabla J_\beta(\theta) &= \EE{\tau}{\sum_{t=0}^\infty\nabla_{\theta}\log\pi_{\theta}(a_{t} \mid s_{t}) \cdot \frac{1}{\beta} e^{\beta\sum_{t'=t}^\infty\gamma^{t'} r(s_{t'}, a_{t'})}} \nonumber \\
&= \EE{\tau}{\sum_{t=0}^\infty\nabla_{\theta}\log\pi_{\theta}(a_{t} \mid s_{t}) \cdot \frac{1}{\beta} e^{\beta R(t)}},
\end{align}

where $ R(t) \defeq \sum_{t'=t}^\infty\gamma^{t'} r(s_{t'}, a_{t'})$ is the discounted rewards-to-go as we defined in Section \ref{sec:reinforce}. Then we can get an empirical estimation of the gradient with $N$ trajectories and horizon $H$ in practice: $\hnabla J_\beta(\theta) = \frac{1}{N}\sum_{i=1}^N\sum_{t=0}^{H-1}\nabla_{\theta}\log\pi_{\theta}(a_{t}^i \mid s_{t}^i) \cdot \frac{1}{\beta} e^{\beta R^i(t)}$.

The risk-sensitive REINFORCE algorithm updates the policy parameters with gradient ascent: $\theta_{t+1}  =  \theta_t + \eta_t\hnabla J_\beta(\theta_t)$, where $\eta_t > 0$ is the learning rate at the $t$-th iteration.

Building on the Lipschitz smoothness assumption and following \cite{khaled2020better}, we introduce a new assumption that bounds the second moment of the gradient for the risk-sensitive objective. This assumption is essential for analyzing convergence properties of the risk-sensitive algorithm. A more detailed discussion of this assumption can be found in \cite{khaled2020better}, where it is used to establish convergence guarantees for non-convex optimization.

\begin{assumption}[Expected smoothness] \label{ass:abc}
    There exists constants $A, B, C \in \mathbb{R}_{\geq 0}$ such that for all $\theta \in \sR^d$, the policy gradient estimator $\widehat{\nabla} J_\beta (\theta)$ satisfies: 
    \begin{equation}
        \E{\norm{\widehat{\nabla} J_\beta (\theta)}^2} \leq 2A(J_\beta^*-J_\beta(\theta))+B\norm{\nabla J_\beta(\theta)}^2+C. 
    \end{equation}
\end{assumption} 

\section{Iteration Complexity Comparison}
\subsection{Empirical Study} \label{empi}
We conduct a series of experiments on the CartPole \cite{1606.01540} environment to empirically evaluate the learning performance and convergence efficiency of the risk-neutral and risk-sensitive REINFORCE algorithms. CartPole is an RL environment where the goal is to balance a pole on a moving cart by applying left or right forces. The agent observes the cart’s position, velocity, pole angle, and angular velocity, taking actions to keep the pole upright. A reward of +1 is given for each timestep the pole remains balanced. 

Specifically, we examine risk-averse scenarios with different values of $\beta = \{-0.01, -0.1, -1, -10\}$ and compare them against the risk-neutral case as a baseline. We employ an empirical estimation of the gradient by sampling a set of $N=10$ truncated trajectories for each iteration, denoted as $\tau_i = \left(s_0^i, a_0^i, r_0^i, s_1^i, \cdots, s_{H-1}^i, a_{H-1}^i, r_{H-1}^i\right)$, with a horizon of $H=200$. We train the agent for $2000$ episodes with a discount factor $\gamma=0.99$. We use a Multilayer Perceptron (MLP) with three hidden layers as the policy network and Adam \cite{kingma2014adam} as the optimizer with a learning rate of $0.001$.

\begin{figure}[t]
    \centering
    \includegraphics[width=0.95\linewidth]{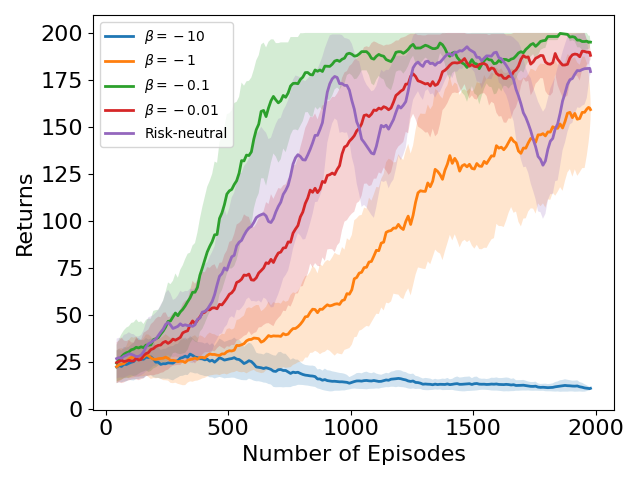}
    \caption{\textbf{Learning curves for risk-neutral and risk-averse algorithms with varying $\beta$ values for the CartPole environment}. The shaded area indicates standard deviation over 10 runs. When $\beta=-0.1$, the learning curve of the risk-averse REINFORCE converges faster and achieves higher returns compared to the risk-neutral REINFORCE.}
    \label{fig:cartpole}
    \vspace{-15pt}
\end{figure}

The results are presented in Figure \ref{fig:cartpole}. We observe that the reward learning curve of the risk-neutral algorithm exhibits instability, whereas introducing risk aversion with $\beta=-0.01$ leads to a more stable learning curve, indicating fewer iterations are required for convergence to a satisfactory solution. Furthermore, when $\beta=-0.1$, the learning curve converges faster, demonstrating an improved acceleration effect and achieving a higher return. These empirical findings confirm that risk-sensitive REINFORCE can achieve more efficiency compared to its risk-neutral counterpart, and also higher return simultaneously. A similar trend, where risk-sensitive REINFORCE converges faster and yields improved performance, was also observed in \cite{noorani2021risk}.

Given these empirical observations, we now aim to rigorously analyze the underpinnings of this phenomenon. Specifically, we seek to address the following questions: (1) \textbf{Formal Characterization and Understanding}: \textit{How can we formally characterize and understand the improved convergence properties of the risk-sensitive REINFORCE}? (2) \textbf{Parameter Conditions}: \textit{Under what conditions can this improved convergence be achieved, especially with regard to the risk-sensitive parameter}? 


\subsection{Risk-Sensitive REINFORCE Iteration Complexity} \label{sec:risk sc}
We first establish the iteration complexity analysis for the risk-sensitive REINFORCE algorithm with the objective of reaching a first-order stationary point (FOSP), leveraging Assumption \ref{ass:abc}, which bounds the second moment of the gradient. If Assumptions \ref{ass: smooth} and \ref{ass:abc} hold, we derive the following corollary: 
 \vspace{8pt}
\begin{coro} \label{cor:abc}
For the risk-sensitive objective $J_\beta$, the Lipschitz smoothness constant is denoted as $L_\beta$. The stepsize $\eta$ falls within the range of $\left(0, \frac{2}{LB}\right)$. Here, we note that $B=0$ implies
that $\eta \in (0, \infty) $.
Define $\delta_0 \defeq J_\beta^* - J_\beta(\theta_0)$. We have that:
\vspace{-.1in}
\begin{align}
\min_{0\leq t\leq T-1} & \E{\norm{\nabla J_\beta(\theta_t)}^2} \leq \frac{2\delta_0(1+L_\beta \eta^2A)^T}{\eta T(2-L_\beta B\eta)} + \frac{L_\beta C\eta}{2-L_\beta B\eta}.\nonumber
\end{align}
\end{coro}

\vspace{8pt}
Please refer to the proof of Corollary \ref{cor:abc} in Appendix\footnote{\url{https://anonymous.4open.science/w/riskrl/}}. We then derive the following Corollary \ref{cor:sc}, whose proof can be found in \citet{khaled2020better}.
\vspace{8pt}

\begin{coro} \label{cor:sc}
In the context of Corollary \ref{cor:abc}, and for a given  $\epsilon > 0$, let $\eta = \min\big\{\frac{1}{\sqrt{L_\beta AT}}, \frac{1}{L_\beta B}, \frac{\epsilon}{2L_\beta C}\big\}$. If the number of iterations $T$ satisfies the following condition: 
\begin{eqnarray} \label{eq:n_beta}
    T \geq \frac{12\delta_0L_\beta}{\epsilon^2}\max\left\{B, \frac{12\delta_0A}{\epsilon^2}, \frac{2C}{\epsilon^2}\right\} \defeq n_\beta, \label{eq:T}
\end{eqnarray}
then $\displaystyle \min_{0\leq t\leq T-1} \E{\norm{\nabla J_\beta(\theta_t)}} \leq \epsilon$. 
\end{coro}

\vspace{8pt}
According to Corollary \ref{cor:abc} and Corollary \ref{cor:sc}, the iteration complexity of obtaining the full-gradient of the risk-sensitive REINFORCE algorithm is $T = \cO(\epsilon^{-2})$. After at least $n_\beta$ iterations, we can achieve an $\epsilon$-approximate FOSP. The complexity depends on the Lipschitz smoothness constant $L_\beta$.

\subsection{Characterizing the Relative Performance} \label{sec:ana}
As in Equation \ref{eq:n_beta}, we define the number of iterations to achieve an $\epsilon$-approximate FOSP for risk-neutral REINFORCE as follows:
\begin{equation} \label{eq:n}
    n \defeq \frac{12\delta_0L}{\epsilon^2}\max\left\{B, \frac{12\delta_0A}{\epsilon^2}, \frac{2C}{\epsilon^2}\right\},
\end{equation}
where $L$ is the Lipschitz smoothness constant for the risk-neutral objective $J$.

Upon analyzing the iteration complexity of the risk-sensitive REINFORCE in Section \ref{sec:risk sc} and the risk-neutral REINFORCE in Equation \ref{eq:n}, as well as empirically evaluating their learning performance in Section \ref{empi}, we conduct a comparison analysis of their iteration complexities. This analysis aims to provide a deeper understanding of their differences in convergence behavior, characterizing their relative performance, and addressing the two key questions raised in Section \ref{empi}.

To address Question (1) regarding the formal characterization and understanding of risk-sensitive REINFORCE's improved convergence, we proceed by comparing $n$ and $n_\beta$, which represent the minimum number of iterations required to reach an $\epsilon$-approximate FOSP for risk-neutral and risk-sensitive REINFORCE, respectively. If we can make $n_\beta$ smaller than $n$, we expect that risk-sensitive REINFORCE may offer improved efficiency, as iteration complexity is expressed as a lower bound on the number of iterations required for convergence. Achieving this reduction would be highly beneficial, as it implies that when considering risk and robustness during the decision-making process, we can have more efficient learning algorithms simultaneously. To formalize our analysis, we assume there exists values of risk-sensitive parameter $\beta$, such that the following holds:
\begin{equation} \label{eq: risk-value-ratio}
    \frac{1}{|\beta|}  e^{|\beta| \sum_{t=0}^\infty\gamma^t\left|r(s_t, a_t)\right|} < \sum_{t=0}^\infty\gamma^t|r(s_t, a_t)|, 
\end{equation}  
Let's define $\alpha$ as the ratio between these two values:
\begin{equation}\label{eq: ratio}
\alpha = \frac{ e^{|\beta|\sum_{t=0}^\infty\gamma^t|r(s_t,a_t)|}}{|\beta| \sum_{t=0}^\infty\gamma^t|r(s_t,a_t)|}, \ 0<\alpha<1.
\end{equation}
\vspace{5pt}

Equation \ref{eq: risk-value-ratio} essentially implies that we aim to identify values of the risk-sensitivity parameter $\beta$, such that the value function of the risk-sensitive algorithm is smaller than that of the risk-neutral algorithm, along any trajectory $\tau$ comprising sequences of states and actions. It is crucial to emphasize that we seek the existence of such $\beta$ values rather than universally applicable ones, as achieving universality is impractical. Such existence may enable us to attain robustness and faster convergence concurrently. 

Then we present the following common assumption \cite{yuan2022general, khaled2020better} that bounds the gradient and Hessian of the policy, which will be used to derive the Lipschitz smoothness constant $L$ and $L_\beta$ for both risk-neutral and risk-sensitive REINFORCE algorithms.

\begin{assumption} \label{ass:lipschitz_smooth_policy}
There exists constants $F_1, F_2 > 0$ such that for every state $s\in S$, the expected gradient and Hessian of $\log\pi_\theta(\cdot\mid s)$ satisfy 
\begin{align}
\EE{a\sim\pi_\theta(\cdot\mid s)}{\norm{\nabla_\theta\log\pi_\theta(a \mid s)}^2} & \leq F_1^2, \label{eq:G2} \\
\EE{a\sim\pi_\theta(\cdot\mid s)}{\norm{\nabla^2_\theta\log\pi_\theta(a\mid s)}} &\leq F_2. \label{eq:F} 
\end{align}
\end{assumption}
\vspace{5pt}

The Lipschitz smoothness constant quantifies the smoothness of a function, particularly its gradient. While a smaller Lipschitz smoothness constant between two algorithms does not necessarily ensure faster convergence, as it merely acts as an upper bound on the gradient, our analysis, as outlined in Corollary \ref{cor:sc}, suggests that the number of iterations required for convergence linearly depends on the Lipschitz smoothness constant. Thus, reducing the Lipschitz smoothness can effectively decrease the number of iterations, which implies better iteration complexity when comparing two algorithms if they have the same computation per iteration. This holds true for our comparison between risk-sensitive and risk-neutral REINFORCE, as employing exponential utility does not introduce additional computation cost.

\begin{theorem} \label{the: Lip}
Suppose that Assumptions \ref{ass: smooth} and \ref{ass:lipschitz_smooth_policy} hold, we have that
\begin{eqnarray}
    L &=& \frac{r_{max}}{(1-\gamma)^2}(F_1^2 + F_2), \nonumber \\
    L_\beta &=& \alpha L,
\end{eqnarray}
\end{theorem}
where $0 < \alpha < 1$ is the ratio in Equation \ref{eq: risk-value-ratio}, representing a multiplication factor that reduces $L_\beta$ compared to $L$. 

The degree of reduction depends on the specific value of the risk-sensitivity parameter $\beta$. $r_{max}$ is the maximum reward, $\gamma$ is the discount factor, $F_1$ and $F_2$ are the constants that bound the gradient and the Hessian of the policy as indicated in Assumption \ref{ass:lipschitz_smooth_policy}. We give the proof of derivation for $L$ and $L_\beta$ in Appendix\footnote{\url{https://anonymous.4open.science/w/riskrl/}}. 

As stated in Equation \ref{eq: ratio}, let's define $x=\sum_{t=0}^\infty\gamma^t|r(s_t,a_t)|$, therefore: $\alpha(x) = \frac{1}{|\beta|}\frac{e^{|\beta|x}}{x}, \ x>0$. While the value function is highly related to the policy gradient procedure, here we can treat it as a variable and compute the first order derivative of $\alpha$ over it. Whether $\alpha$ can take the optimum value $x^*$ achieved by setting the first-order derivative to zero, will depend on the policy gradient procedure. 

Take the first order derivative of $\alpha$ w.r.t. $x$, we have $\nabla \alpha(x) = \frac{1}{|\beta|} \frac{e^{|\beta|x}(|\beta|x-1)}{x^2}$, when $\nabla \alpha(x^*) = 0, x^* = \frac{1}{|\beta|}$. If $\alpha$ can take the minimum value at $x^*$, then $\alpha_{min}=\beta^2 e$. In order to have $0<\alpha<1$ for the existence of some values of $\beta$, we need $\beta^2 e < 1$, then $|\beta| < e^{-\frac{1}{2}}$. Furthermore, $x=\sum_{t=0}^\infty\gamma^t|r(s_t,a_t)| \leq \frac{r_{max}}{1-\gamma}$, therefore, $\frac{1}{|\beta|} < \frac{r_{max}}{1-\gamma}$, then, $|\beta| > \frac{1-\gamma}{r_{max}}$.

Based on the above analysis, we have the following remark that there exists a range of values for $\beta$:
\begin{equation} \label{eq:beta_range}
    \frac{1-\gamma}{r_{max}} < |\beta| < e^{-\frac{1}{2}},
\end{equation}
leading to $n_\beta < n$, which potentially implies a more efficient risk-sensitive REINFORCE compared to the risk-neutral REINFORCE in terms of the iteration complexity. This addresses Question (2) regarding the risk-sensitive parameter. However, we clarify that this result demonstrates the existence of a range of values for $\beta$; it does not imply that values outside this range are necessarily suboptimal. 

\section{Experiments}
To empirically validate our analysis in Section \ref{sec:ana}, we conduct experiments across multiple environments, including CartPole \cite{1606.01540}, holonomic robot navigation, and Minigrid \cite{MinigridMiniworld23}. As shown in Figure \ref{fig:cartpole}, the CartPole experiments demonstrate that the risk-sensitive REINFORCE can achieve greater efficiency compared to its risk-neutral counterpart while also attaining higher returns when $\beta=-0.1$.

\subsection{Holonomic Robot Navigation}
\label{subsec:holonomic_experiment}
Beyond the CartPole environment, we extend our experiments into Holonomic Robot Navigation, a continuous and dynamic environment. In this setting, we have a mobile robot with holonomic kinematics motion model navigating a 2D workspace with obstacles, aiming to reach a goal.

\begin{figure}[t]
    \centering

     \centering
     \begin{subfigure}[b]{0.49\linewidth}
         \centering
         \includegraphics[width=\textwidth]{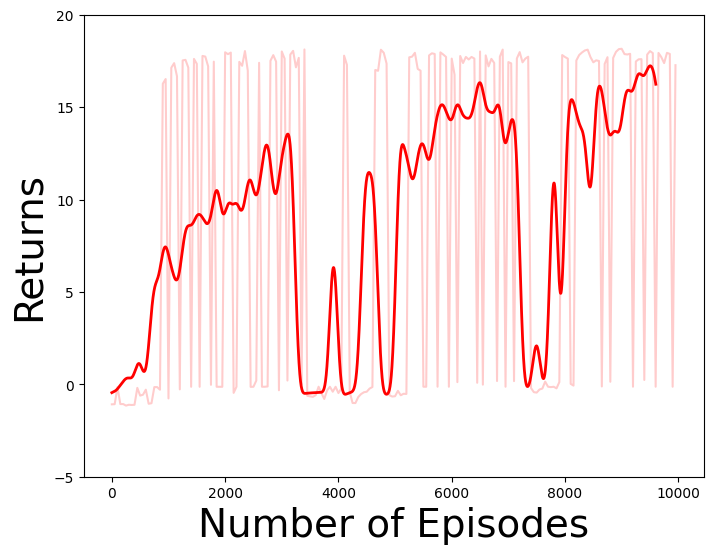}
         \caption{Risk-neutral}
         \label{fig:hr0}
     \end{subfigure}
     \begin{subfigure}[b]{0.49\linewidth}
         \centering
         \includegraphics[width=\textwidth]{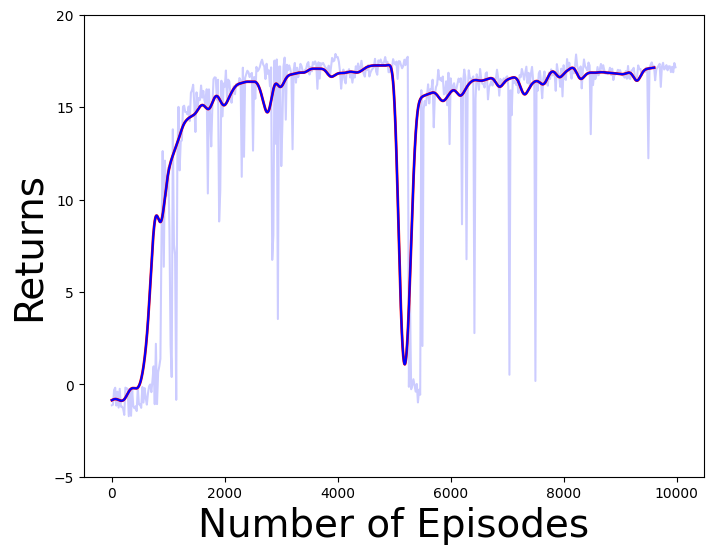}
         \caption{$\beta=-0.5$}
         \label{fig:hr1}
     \end{subfigure}
     \vfill
     \begin{subfigure}[b]{0.49\linewidth}
         \centering
         \includegraphics[width=\textwidth]{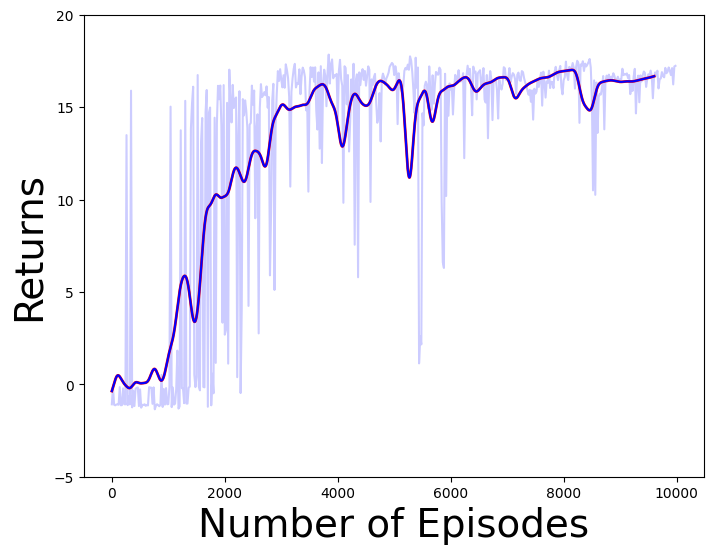}
         \caption{$\beta=-1.0$}
         \label{fig:hr3}
     \end{subfigure}
     \begin{subfigure}[b]{0.49\linewidth}
         \centering
         \includegraphics[width=\textwidth]{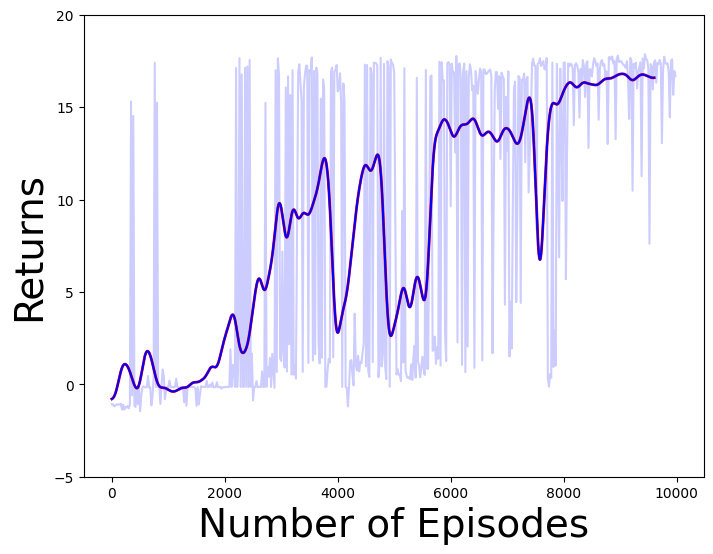}
         \caption{$\beta=-5.0$}
         \label{fig:hr4}
     \end{subfigure}
    \caption{\textbf{Learning curves for risk-neutral and risk-averse  cases with varying $\beta$ values in the holonomic robot navigation environment}. The solid lines represent average returns over 10 runs, while the shaded lines indicate returns from an individual run. The risk-neutral policy exhibits large deviations and instability, with excessive oscillations. In contrast, the risk-averse policies with $\beta = -0.5$ and $\beta = -1.0$ demonstrate greater stability, faster convergence, and higher returns. However, when $\beta=-5.0$, the learning efficiency decreases because an excessively large magnitude of $\beta$ leads to an overly conservative policy that prioritizes obstacle avoidance to the extent that learning efficiency is compromised.}
    \label{fig:lr_h}
    \vspace{-15pt}
\end{figure}

\begin{figure}[ht]
    \centering
    \includegraphics[width=0.83
    \linewidth]{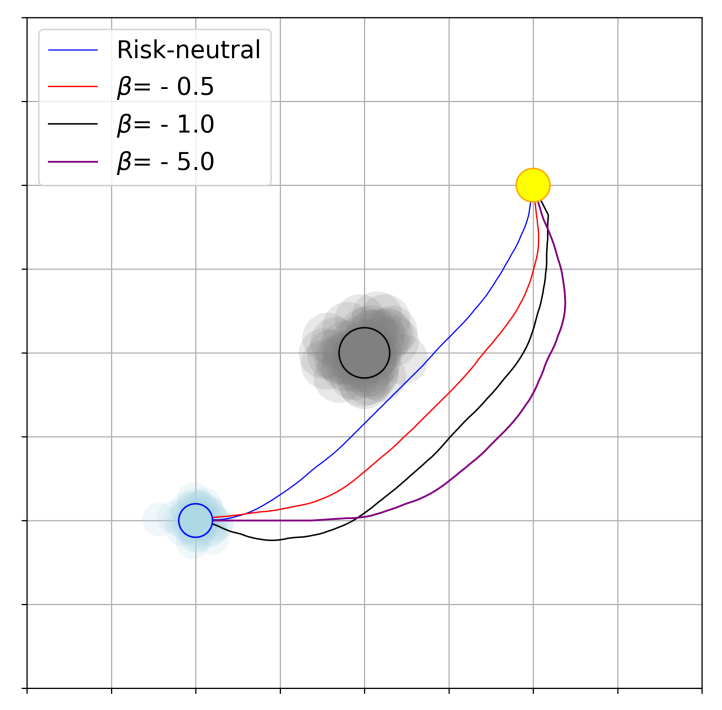}
    \caption{\textbf{Sample navigation trajectories comparing risk-neutral and risk-averse policies with varying $\beta$ values in the holonomic robot navigation environment.} The light blue dot represents the starting position, the yellow dot indicates the goal, and the gray dot in between represents the obstacle. The risk-neutral policy exhibits aggressive movements, while the risk-averse policies follow more stable and conservative paths.}
    \label{fig:holonomic_trajectories}
    \vspace{-15pt}
\end{figure}

Unlike traditional differential drive robots, holonomic robots can move in any direction instantaneously, allowing full control over velocity in both the \textit{x} and \textit{y} directions. The robot's state at time step $t$ is represented as $s_t = [x_t, y_t, v_{x,t}, v_{y,t}, o_1, o_2, r_o]$, where $(x_t, y_t)$ denotes the agent's position, $(v_{x,t}, v_{y,t})$ is its velocity, and $(o_1, o_2, r_o)$ denote the obstacle positions and their radii. The action space consists of discrete velocity increments: $a_t \in \{(\Delta v_x, \Delta v_y) \mid \Delta v_x, \Delta v_y \in [-0.1, 0, 0.1] \}$, which updates the robot's velocity. The reward function is distance-based and encourages the agent to reach the goal efficiently: $r_t = (\frac{50}{d_t} - d_t) \times 10^{-4}$, where $d_t$ is the Euclidean distance to the goal. 




We train the agent for 10000 episodes, sampling $N=10$ trajectories per iteration with a horizon of $H=500$, following the same training setup as in the CartPole environment. We conduct experiments for both the risk-neutral and risk-averse scenarios, evaluating different values of $\beta = \{-0.5, -1.0, -5.0\}$, with the risk-neutral case serving as a baseline for comparison.



Figure \ref{fig:lr_h} presents the learning curves for risk-neutral and risk-averse cases with different values of $\beta$. The risk-neutral policy exhibits large deviations and instability, with excessive oscillations. In contrast, the risk-averse policies with $\beta = -0.5$ and $\beta = -1.0$ demonstrate greater stability, faster convergence, and higher returns. However, when the magnitude of $\beta$ becomes overly large, as shown in Figure \ref{fig:hr4} for $\beta = -5.0$, the policy requires more training episodes to reach optimal performance. This occurs because an overly conservative policy prioritizes obstacle avoidance to the extent that learning efficiency decreases.

To further analyze policy behaviors, we visualize sampled trajectories for the converged policies in Figure~\ref{fig:holonomic_trajectories}. The risk-neutral policy exhibits aggressive movements, while the risk-averse policies follow more stable and conservative paths, reducing collisions as the magnitude of $\beta$ increases. Additionally, inherent randomness in the policy can lead to potential collisions, prompting risk-averse agents to maintain greater distances from obstacles. This risk-averse behavior is particularly beneficial in scenarios where safety is a priority, especially under uncertainty in ego and obstacle positions due to perception noise and dynamic variations. Moreover, the experimental results demonstrate that, beyond enhancing robustness and safety, risk-sensitive policies can also improve learning efficiency.

\subsection{Minigrid}
We conduct additional evaluations in the MiniGrid navigation \cite{MinigridMiniworld23} environment, where the agent's objective is to reach a goal position. Specifically, we utilize the \textit{MiniGrid-Empty-Random-6$\times$6} environment, where the agent is initialized at a random starting position in each episode and receives a sparse reward after successfully reaching the goal. This randomness increases variance in the learning process, making it a suitable testbed for evaluating the algorithms' stability and convergence efficiency.

\begin{figure}[t]
     \centering
     \begin{subfigure}[b]{0.49\linewidth}
         \centering
         \includegraphics[width=\textwidth]{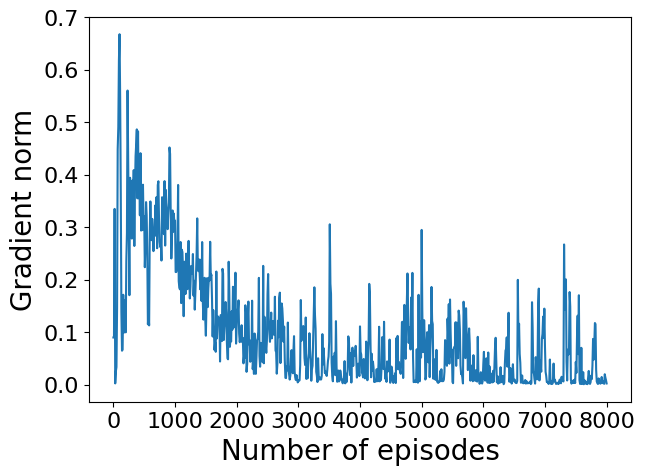}
         \caption{Risk-neutral}
         \label{fig:g0}
     \end{subfigure}
     \hfill
     \begin{subfigure}[b]{0.49\linewidth}
         \centering
         \includegraphics[width=\textwidth]{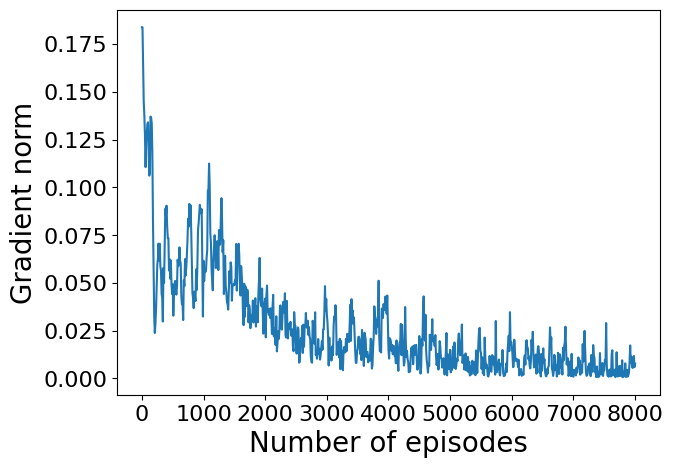}
         \caption{$\beta=-0.1$}
         \label{fig:g1}
     \end{subfigure}
     \vfill
     \begin{subfigure}[b]{0.49\linewidth}
         \centering
         \includegraphics[width=\textwidth]{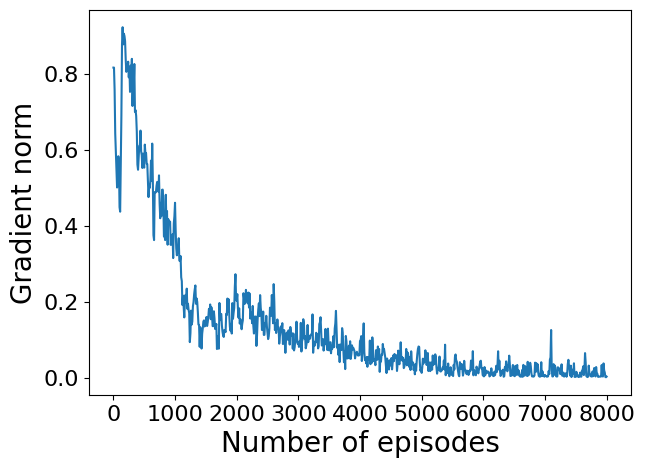}
         \caption{$\beta=-0.5$}
         \label{fig:g3}
     \end{subfigure}
     \begin{subfigure}[b]{0.49\linewidth}
         \centering
         \includegraphics[width=\textwidth]{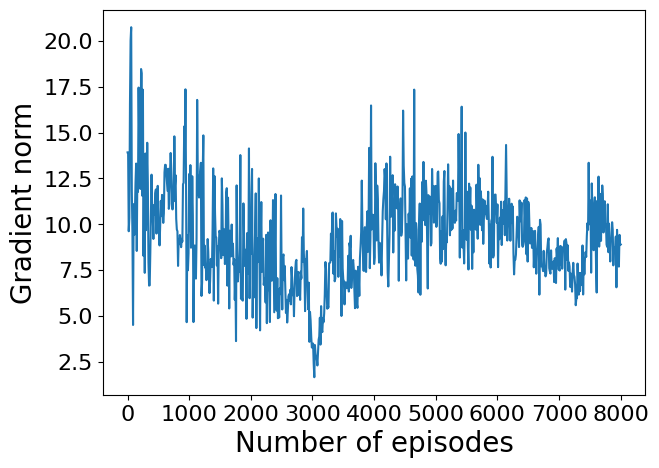}
         \caption{ $\beta=-10.0$}
         \label{fig:g4}
     \end{subfigure}
        \caption{\textbf{The gradient norm of risk-neutral and risk-averse algorithms with varying $\beta$ values for the Minigrid environment}. The gradient norm decreases more rapidly when $\beta=-0.1$ and $\beta=-0.5$ for the risk-averse algorithm compared to its risk-neutral counterpart, and the norm exhibits smoother behavior in risk-averse cases. However, when the magnitude of $\beta$ becomes overly large ($\beta = -10.0$), the gradient norm becomes large, impeding the learning process.}
        \label{fig:g}
        \vspace{-10pt}
\end{figure}

\begin{figure}[t]
     \centering
     \begin{subfigure}[b]{0.49\linewidth}
         \centering
         \includegraphics[width=\textwidth]{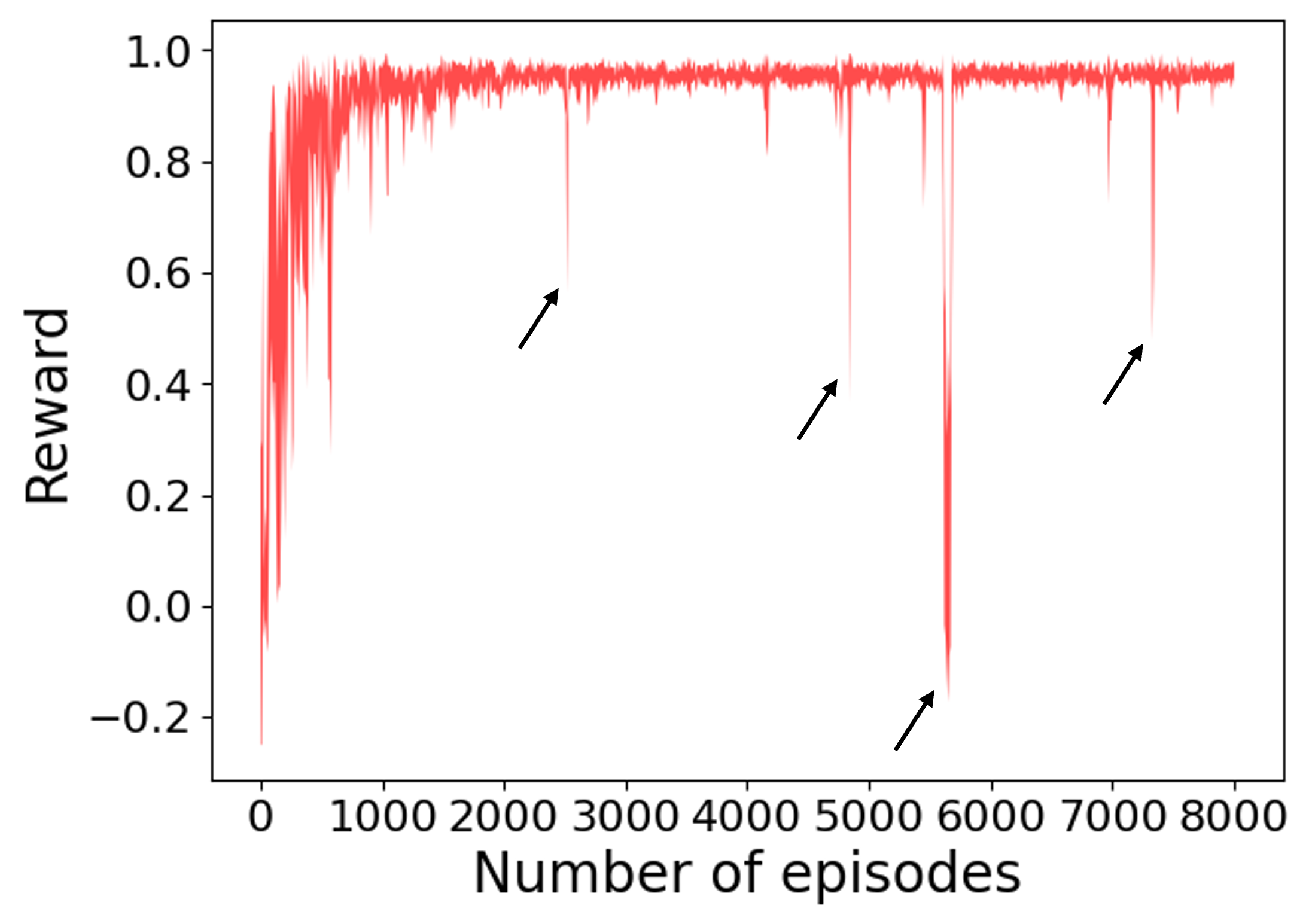}
         \caption{Risk-neutral}
         \label{fig:r0}
     \end{subfigure}
     \begin{subfigure}[b]{0.49\linewidth}
         \centering
         \includegraphics[width=\textwidth]{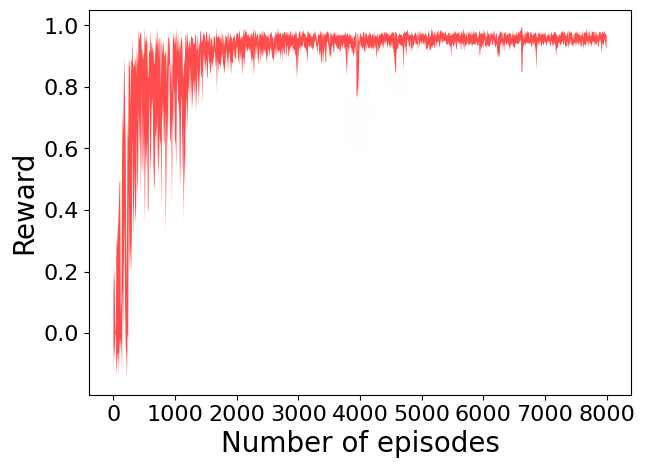}
         \caption{$\beta=-0.1$}
         \label{fig:r1}
     \end{subfigure}
     \vfill
     \begin{subfigure}[b]{0.49\linewidth}
         \centering
         \includegraphics[width=\textwidth]{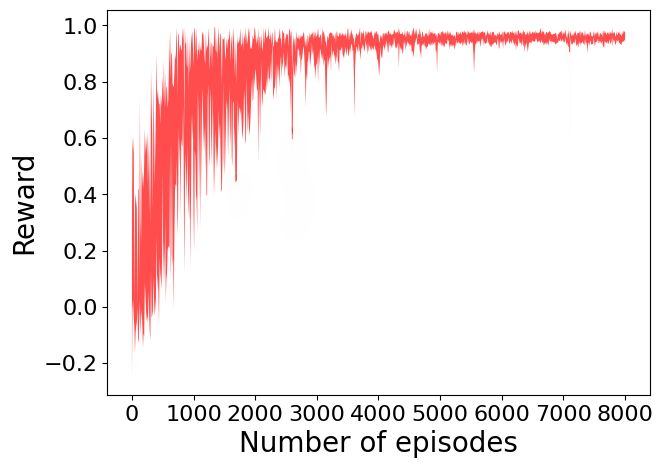}
         \caption{$\beta=-0.5$}
         \label{fig:r3}
     \end{subfigure}
     \begin{subfigure}[b]{0.49\linewidth}
         \centering
         \includegraphics[width=\textwidth]{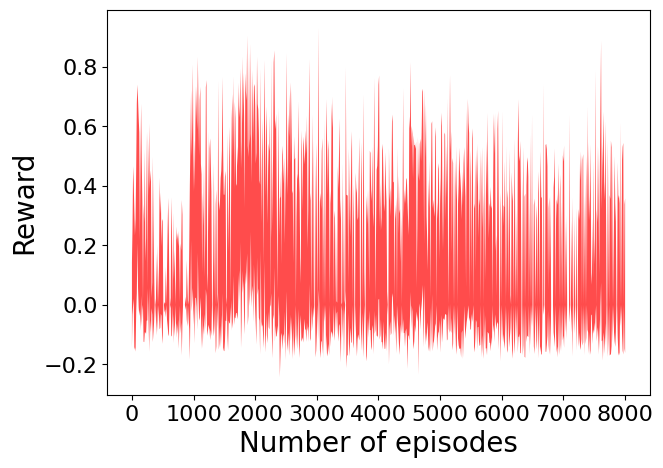}
         \caption{$\beta=-10.0$}
         \label{fig:r4}
     \end{subfigure}
        \caption{\textbf{Learning curves for risk-neutral and risk-averse algorithms with varying $\beta$ values for the Minigrid environment}. Arrows ($\uparrow$) in (a) depict extreme values. The risk-neutral case displays more variability and more significant extreme values. In contrast, the risk-averse cases exhibit less variability and fewer extreme values when $\beta=-0.1$ and $\beta=-0.5$, suggesting that they require fewer episodes to converge and stabilize. However, when $\beta=-10.0$, the gradient norm becomes large, leading to an oscillatory learning curve that hinders the learning process.}
        \label{fig:r}
        \vspace{-10pt}
\end{figure}

We employ an empirical estimation of the gradient by sampling a set of $N=10$ truncated trajectories for each iteration with a horizon of $H=200$. We train the agent for $8000$ episodes with the same training setup as used in the CartPole environment. We conduct experiments for both the risk-neutral and risk-averse scenarios with different values of $\beta=\{-0.1, -0.5, -10.0\}$, utilizing the risk-neutral case as a baseline for comparison. 


In Figure \ref{fig:g}, we depict the gradient norm $\norm{\nabla J}$ for the risk-neutral case and $\norm{\nabla J_\beta}$ for risk-averse cases with different values of $\beta$. The results indicate that the gradient norm decreases more rapidly for the risk-averse algorithm when $\beta=-0.1$ and $\beta=-0.5$ compared to its risk-neutral counterpart, exhibiting smoother behavior across iterations. Based on the $\epsilon$-approximate FOSP convergence criterion, which requires the gradient norm to be at most $\epsilon$, these observations suggest that the risk-averse algorithm achieves convergence with fewer iterations compared to its risk-neutral counterpart, under appropriate risk-sensitive parameters $\beta$. However, when $\beta = -10.0$, an overly large magnitude of $\beta$ leads to a large gradient norm, impeding the learning process.

In Figure \ref{fig:r}, we illustrate the learning curves for the risk-neutral case and for different values of $\beta$ in the risk-averse cases. While the acceleration effects for the risk-averse cases are not pronounced compared to the risk-neutral case, the risk-averse cases with $\beta=-0.1$ and $\beta=-0.5$ exhibit less variability and fewer extreme values, indicating that they require fewer episodes to converge and stabilize. In contrast, the risk-neutral case displays more variability and more significant extreme values, as highlighted by arrows in Figure \ref{fig:r0}. This observation verifies that risk-averse algorithms can potentially achieve reduced iteration complexity, which also aligns with Figure \ref{fig:g}, illustrating that the risk-averse algorithm converges faster to a FOSP.


\section{Conclusions}
In conclusion, we conduct a comprehensive analysis of the iteration complexity of the risk-sensitive REINFORCE algorithm, achieving an iteration complexity of $\cO(\epsilon^{-2})$ aimed at attaining a First-Order Stationary Point (FOSP). This represents the first extension of iteration complexity analysis to the risk-sensitive REINFORCE algorithm. We further compare the iteration complexity of the risk-sensitive REINFORCE against its risk-neutral counterpart, which serves as a baseline. Our findings indicate that the risk-sensitive algorithm can potentially achieve convergence with fewer iterations. This finding is significant as it suggests that while considering safety during the decision-making process, we can simultaneously make the learning algorithm more efficient. To validate our analysis, we conduct experiments across multiple environments, including CartPole, MiniGrid, and Holonomic Robot Navigation, evaluating performance under different risk-sensitive parameters. Empirical results confirm that risk-sensitive policies can not only converge faster but also exhibit more stable learning behavior compared to their risk-neutral counterparts, aligning with our findings.

\textbf{Limitations and Future Work.} While our work offers valuable insights, it has certain limitations. We focus on the risk-sensitive REINFORCE algorithm with an exponential utility function to integrate safety into the learning objective. Future research could explore alternative risk measures, and extend the analysis to other policy gradient methods beyond REINFORCE.

\bibliography{references}


%
%





%

%


\onecolumn
\begin{center}
\textbf{\huge Appendix}
\end{center}

\section{Proof of Corollary \ref{cor:abc}} \label{apx: abc}
We initiate our analysis by considering Assumption ~\ref{ass: smooth}, which concerns Lipschitz smoothness. For the risk-sensitive objective $J_\beta$ with a Lipschitz smoothness constant $L_\beta$, we have that: 
\begin{eqnarray}
    J_\beta(\theta_{t+1}) &\geq& J_\beta(\theta_t) + \dotprod{\nabla J_\beta(\theta_t), \theta_{t+1} - \theta_t} - \frac{L_\beta}{2}\norm{\theta_{t+1}-\theta_t}^2 \nonumber \\
    &=& J_\beta(\theta_t) + \eta\dotprod{\nabla J_\beta(\theta_t), \widehat{\nabla} J_\beta(\theta_t)} - \frac{L_\beta\eta^2}{2}\norm{\widehat{\nabla} J_\beta(\theta_t)}^2.
\end{eqnarray}
Take expectations on both sides conditioned on $\theta_t$ and use Assumption \ref{ass:abc}, we get: 
\begin{eqnarray}
\EE{t}{J_\beta(\theta_{t+1})} &\geq& J_\beta(\theta_t) + \eta\dotprod{\nabla J_\beta(\theta_t), \nabla J_\beta(\theta_t)} - \frac{L_\beta\eta^2}{2}\EE{t}{\norm{\widehat{\nabla} J_\beta(\theta_t)}^2} \nonumber \\
&\geq& J_\beta(\theta_t) + \eta\norm{\nabla J_\beta(\theta_t)}^2 - \frac{L_\beta\eta^2}{2}\left(2A(J_\beta^*-J_\beta(\theta_t)) + B\norm{\nabla J_\beta(\theta_t)}^2 + C\right) \nonumber \\
&=& J_\beta(\theta_t) + \eta\left(1-\frac{L_\beta B\eta}{2}\right)\norm{\nabla J_\beta(\theta_t)}^2 - L_\beta\eta^2A(J_\beta^*-J_\beta(\theta_t)) - \frac{L_\beta C\eta^2}{2}. \nonumber \\
\end{eqnarray}
Then we subtract $J_\beta^*$ from both sides,
\begin{eqnarray}
\EE{t}{J_\beta(\theta_{t+1})} - J_\beta^* &\geq& -(1+L_\beta \eta^2A)(J_\beta^* - J_\beta(\theta_t)) + \eta\left(1-\frac{L_\beta B\eta}{2}\right)\norm{\nabla J_\beta(\theta_t)}^2 - \frac{L_\beta C\eta^2}{2}.
\end{eqnarray}
Take the expectation on both sides and rearrange the equation, we obtain:
\begin{eqnarray}
\E{J_\beta^* - J_\beta(\theta_{t+1})} + \eta\left(1-\frac{L_\beta B\eta}{2}\right)\E{\norm{\nabla J_\beta(\theta_t)}^2} \leq (1+L_\beta \eta^2A)\E{J_\beta^*-J_\beta(\theta_t)} + \frac{L_\beta C\eta^2}{2}.
\end{eqnarray}
Define $\delta_t \defeq \E{J_\beta^* - J_\beta(\theta_t)}$ and $r_t \defeq \E{\norm{\nabla J_\beta(\theta_t)}^2}$, we can rewrite the above inequality as
\begin{eqnarray} \label{eq:introduce_weight}
\eta\left(1-\frac{L_\beta B\eta}{2}\right)r_t &\leq& (1+L_\beta \eta^2A)\delta_t - \delta_{t+1} + \frac{L_\beta C\eta^2}{2}.
\end{eqnarray}
Now, we introduce a sequence of weights, denoted as $w_{-1}, w_0, w_1, \cdots, w_{T-1}$, based on a method used by \cite{stich2019unified, khaled2020better, yuan2022general}. We initialize $w_{-1}$ with a positive value. We define $w_t$ as $w_t \eqdef \frac{w_{t-1}}{1+L_\beta \eta^2A}$ for all $t\geq0$. It's important to note that when $A=0$, all $w_t$ are equal, i.e., $w_t=w_{t-1}=\cdots=w_{-1}$. By multiplying~\eqref{eq:introduce_weight} by $w_t/\eta$, we can derive:
\begin{eqnarray}
\left(1-\frac{L_\beta B\eta}{2}\right)w_tr_t &\leq& \frac{w_t(1+L_\beta \eta^2A)}{\eta}\delta_t - \frac{w_t}{\eta}\delta_{t+1} + \frac{L_\beta C\eta}{2}w_t  \nonumber \\
&=& \frac{w_{t-1}}{\eta}\delta_t - \frac{w_t}{\eta}\delta_{t+1} + \frac{L_\beta C\eta}{2} w_t.
\end{eqnarray}
When we sum up both sides for $t=0,1,\cdots,T-1$, we get:
\begin{eqnarray} \label{eq:wtrt}
\left(1-\frac{L_\beta B\eta}{2}\right)\sum_{t=0}^{T-1}w_tr_t &\leq& \frac{w_{-1}}{\eta}\delta_0 - \frac{w_{T-1}}{\eta}\delta_T + \frac{L_\beta C\eta}{2} \sum_{t=0}^{T-1}w_t \nonumber \\
&\leq& \frac{w_{-1}}{\eta}\delta_0 + \frac{L_\beta C\eta}{2}\sum_{t=0}^{T-1}w_t.
\end{eqnarray}
We can define $W_T$ as $W_T \eqdef \sum_{t=0}^{T-1}w_t$. By dividing both sides of the equation by $W_T$, we obtain:
\begin{eqnarray} \label{eq:min_rt}
\left(1-\frac{L_\beta B\eta}{2}\right)\min_{0\leq t\leq T-1}r_t \leq \frac{1}{W_T}\cdot\left(1-\frac{L_\beta B\eta}{2}\right)\sum_{t=0}^{T-1}w_tr_t \leq \frac{w_{-1}}{W_T}\frac{\delta_0}{\eta} + \frac{L_\beta C\eta}{2}.
\end{eqnarray}
Note that,
\begin{eqnarray}
W_T = \sum_{t=0}^{T-1}w_t \geq \sum_{t=0}^{T-1}\min_{0\leq i\leq T-1}w_i = Tw_{T-1} = \frac{Tw_{-1}}{(1+L_\beta \eta^2A)^T}.
\end{eqnarray}
Use this in \eqref{eq:min_rt},
\begin{eqnarray} \label{eq:min_rt2}
\left(1-\frac{L_\beta B\eta}{2}\right)\min_{0\leq t\leq T-1}r_t &\leq& \frac{(1+L\eta^2A)^T}{\eta T}\delta_0 + \frac{LC\eta}{2}.
\end{eqnarray}
By substituting $r_t$ in \eqref{eq:min_rt2} with $\E{\norm{\nabla J_\beta (\theta_t)}^2}$, we obtain: 
\begin{eqnarray}
\left(1-\frac{L_\beta B\eta}{2}\right)\min_{0\leq t\leq T-1}\E{\norm{\nabla J_\beta (\theta_t)}^2} &\leq& \frac{(1+L_\beta \eta^2A)^T}{\eta T}\delta_0 + \frac{L_\beta C\eta}{2}, \nonumber \\
\min_{0\leq t\leq T-1} \E{\norm{\nabla J_\beta(\theta_t)}^2} &\leq& \frac{2\delta_0(1+L_\beta \eta^2A)^T}{\eta T(2-L_\beta B\eta)} + \frac{L_\beta C\eta}{2-L_\beta B\eta}.
\end{eqnarray}

The choice of our step size ensures that for both cases, whether $B > 0$ or $B=0$, we have $1-\frac{L_\beta B\eta}{2} > 0$. 
\section{Proof of Theorem \ref{the: Lip}} \label{apx: comparison}
\begin{lemma} \label{lem:E_tau_nabla}
Subject to Assumption~\ref{ass:lipschitz_smooth_policy}, for any non-negative integer $t$, and for any state-action pair $(s_t, a_t) \in \cS \times \cA$ at time $t$ within a trajectory $\tau$ sampled under the parameterized policy $\pi_\theta$, we have the following:
\begin{eqnarray}
\EE{\tau\sim p(\cdot\mid\theta)}{\norm{\nabla_\theta\log\pi_\theta(a_t \mid s_t)}^2} &\leq& F_1^2, \label{eq:E_tau_F1} \\
\EE{\tau\sim p(\cdot\mid\theta)}{\norm{\nabla^2_\theta\log\pi_\theta(a_t \mid s_t)}} &\leq& F_2. \label{eq:E_tau_F2}
\end{eqnarray}
\end{lemma}

For $t > 0$ and $(s_t, a_t) \in \cS \times \cA$, we have
\begin{align}
\EE{\tau}{\norm{\nabla_{\theta}\log\pi_{\theta}(a_t \mid s_t)}^2} &= \EE{s_t}{\EE{a_t\sim \pi_\theta(\cdot\mid s_t)}{\norm{\nabla_{\theta}\log\pi_{\theta}(a_t \mid s_t)}^2 | s_t}}\overset{\ref{eq:G2}}{\leq} F_1^2,
\end{align}
where the first equality is obtained by the Markov property. Similarly, we have
\begin{align} 
\EE{\tau}{\norm{\nabla_{\theta}^2\log\pi_{\theta}(a_t \mid s_t)}} = \EE{s_t}{\EE{a_t\sim \pi_\theta(\cdot\mid s_t)}{\norm{\nabla_{\theta}^2\log\pi_{\theta}(a_t \mid s_t)} | s_t}}\overset{\ref{eq:F}}{\leq}  F_2.
\end{align}

Then Lemma \ref{lem:E_tau_nabla} is then used for the derivation of $L$ and $L_\beta$.

Assumption \ref{ass: smooth} is equivalent to $\norm{\nabla^2 J(\theta)} \leq L$ for the risk-neutral REINFORCE and $\norm{\nabla^2 J_\beta(\theta)} \leq L_\beta$ for the risk-sensitive REINFORCE. We first take the second order derivative of the risk-neutral objective w.r.t. $\theta$, in order to derive the $L$-Lipschitz smooth constant. 
\begin{eqnarray} \label{eq:neutral-l}
\nabla^2 J(\theta) &\overset{\ref{eq:GD}}{=}& \nabla_\theta\EE{\tau}{\sum_{t=0}^\infty \nabla_{\theta}\log\pi_{\theta}(a_t \mid s_t) R(t)} \nonumber \\
&=& \nabla_\theta \Bigg[\int p(\tau\mid\theta)\sum_{t=0}^\infty \nabla_{\theta}\log\pi_{\theta}(a_t \mid s_t) R(t) d\tau \Bigg] \nonumber \\
&=& \int\nabla_\theta p(\tau\mid\theta)\left(\sum_{t=0}^\infty \nabla_{\theta}\log\pi_{\theta}(a_t \mid s_t) R(t)\right)^\top d\tau + \int p(\tau\mid\theta)\sum_{t=0}^\infty \nabla_{\theta}^2\log\pi_{\theta}(a_t \mid s_t) R(t)d\tau \nonumber \\
&=& \int p(\tau\mid\theta)\nabla_\theta\log p(\tau\mid\theta)\left(\sum_{t=0}^\infty \nabla_{\theta}\log\pi_{\theta}(a_t \mid s_t) R(t)\right)^\top d\tau + \int p(\tau\mid\theta)\sum_{t=0}^\infty \nabla_{\theta}^2\log\pi_{\theta}(a_t \mid s_t) R(t)d\tau \nonumber \\
&=& \EE{\tau}{\nabla_\theta\log p(\tau\mid\theta)\left(\sum_{t=0}^\infty \nabla_{\theta}\log\pi_{\theta}(a_t \mid s_t) R(t)\right)^\top} + \EE{\tau}{\sum_{t=0}^\infty \nabla_{\theta}^2\log\pi_{\theta}(a_t \mid s_t) R(t)} \nonumber \\
&=& \underbrace{\EE{\tau}{\sum_{k=0}^\infty\nabla_\theta\log \pi_\theta(a_k\mid\theta_k)\left(\sum_{t=0}^\infty \nabla_{\theta}\log\pi_{\theta}(a_t \mid s_t) R(t)\right)^\top}}_{\circled{1}} + \underbrace{\EE{\tau}{\sum_{t=0}^\infty \nabla_{\theta}^2\log\pi_{\theta}(a_t \mid s_t) R(t)}}_{\circled{2}}. \label{eq:1+2}
\end{eqnarray}

We individually bound the aforementioned two terms for the risk-neutral REINFORCE.

For the term \circled{1},
\begin{eqnarray}
\norm{\circled{1}} &=& \norm{\EE{\tau}{\sum_{k=0}^\infty\nabla_\theta\log \pi_\theta(a_k\mid\theta_k)\left(\sum_{t=0}^\infty \nabla_{\theta}\log\pi_{\theta}(a_t \mid s_t) R(t)\right)^\top}} \nonumber \\
&=& \norm{\EE{\tau}{\sum_{k=0}^\infty\nabla_\theta\log \pi_\theta(a_k\mid\theta_k)\left(\sum_{t=0}^\infty \nabla_{\theta}\log\pi_{\theta}(a_t \mid s_t) \sum_{t'=t}^\infty \gamma^{t'} r(s_{t'}, a_{t'}) \right)^\top}} \nonumber \\
&=& \norm{\EE{\tau}{\sum_{t=0}^\infty\gamma^t r(s_t, a_t)\left(\sum_{t'=0}^t\nabla_\theta\log\pi_\theta(a_{t'}\mid \theta_{t'})\right)\left(\sum_{k=0}^t\nabla_\theta\log\pi_\theta(a_k\mid \theta_k)\right)^\top}} \nonumber \\
&\leq& \EE{\tau}{\sum_{t=0}^\infty\gamma^t\left|r(s_t, a_t)\right|\norm{\sum_{k=0}^t\nabla_\theta\log \pi_\theta(a_{k}\mid\theta_{k})}^2} \nonumber \\
&\leq& r_{max}\sum_{t=0}^\infty\gamma^t\sum_{k=0}^t\EE{\tau}{\norm{\nabla_\theta\log \pi_\theta(a_{k}\mid\theta_{k})}^2} \nonumber \\
&\overset{\ref{eq:E_tau_F1}}{\leq}& r_{max}F_1^2\sum_{t=0}^\infty\gamma^t(t+1) \nonumber \\
&=& \frac{r_{max}F_1^2}{(1-\gamma)^2},
\end{eqnarray}
where the third line is due to the fact that the future actions do not depend on the past rewards. 

For the term \circled{2}, 
\begin{eqnarray}
\norm{\circled{2}} &=& \norm{\EE{\tau}{\sum_{t=0}^\infty \nabla_{\theta}^2\log\pi_{\theta}(a_t \mid s_t) R(t)}} \nonumber \\ 
 &=& \norm{\EE{\tau}{\sum_{t=0}^\infty \nabla_{\theta}^2\log\pi_{\theta}(a_t \mid s_t) \sum_{t'=t}^\infty \gamma^{t'} r(s_{t'}, a_{t'})}} \nonumber \\
&=& \norm{\EE{\tau} {\sum_{t=0}^\infty\gamma^{t}r(s_{t},a_{t})\left(\sum_{k=0}^t\nabla_{\theta}^2\log\pi_{\theta}(a_k \mid s_k)\right)}} \nonumber \\
&\leq& \EE{\tau} {\sum_{t=0}^\infty\gamma^{t}\left|r(s_{t},a_{t})\right|\left(\sum_{k=0}^t\norm{\nabla_{\theta}^2\log\pi_{\theta}(a_k \mid s_k)}\right)} \nonumber \\
&\leq& r_{\max}\sum_{t=0}^\infty\gamma^t\left(\sum_{k=0}^t\EE{\tau}{\norm{\nabla_{\theta}^2\log\pi_{\theta}(a_k \mid s_k)}}\right) \nonumber \\ 
&\overset{\ref{eq:E_tau_F2}}{\leq}& r_{\max}F_2\sum_{t=0}^\infty\gamma^t(t+1) \nonumber \\
&=& \frac{r_{\max}F_2}{(1-\gamma)^2},
\end{eqnarray}
where the third line is also due to the fact that the future actions do not depend on the past rewards. 

Finally,
\begin{eqnarray}
\norm{\nabla^2J(\theta)} \leq \frac{r_{\max}}{(1-\gamma)^2}(F_1^2+F_2),
\end{eqnarray}
so $L = \frac{r_{\max}}{(1-\gamma)^2}(F_1^2+F_2)$. 

Then we take the second order derivative of the risk-sensitive objective w.r.t. $\theta$, in order to derive the $L_\beta$-Lipschitz smoothness constant.
\begin{eqnarray} \label{eq:risk-l}
\nabla^2 J_\beta(\theta) &\overset{\ref{eq:risk-pg}}{=}& \nabla_\theta\EE{\tau}{\sum_{t=0}^\infty \nabla_{\theta}\log\pi_{\theta}(a_t \mid s_t)\cdot \frac{1}{\beta} e^{\beta R(t)}} \nonumber \\
&=& \nabla_\theta \Bigg[\int p(\tau\mid\theta)\sum_{t=0}^\infty \nabla_{\theta}\log\pi_{\theta}(a_t \mid s_t) \cdot\frac{1}{\beta} e^{\beta R(t)} d\tau \Bigg] \nonumber \\
&=& \int\nabla_\theta p(\tau\mid\theta)\left(\sum_{t=0}^\infty \nabla_{\theta}\log\pi_{\theta}(a_t \mid s_t) \cdot\frac{1}{\beta} e^{\beta R(t)}\right)^\top d\tau + \int p(\tau\mid\theta)\sum_{t=0}^\infty \nabla_{\theta}^2\log\pi_{\theta}(a_t \mid s_t) \cdot \frac{1}{\beta} e^{\beta R(t)} d\tau \nonumber \\
&=& \int p(\tau\mid\theta)\nabla_\theta\log p(\tau\mid\theta)\left(\sum_{t=0}^\infty \nabla_{\theta}\log\pi_{\theta}(a_t \mid s_t) \cdot\frac{1}{\beta} e^{\beta R(t)}\right)^\top d\tau \nonumber \\ 
&\quad& + \int p(\tau\mid\theta)\sum_{t=0}^\infty \nabla_{\theta}^2\log\pi_{\theta}(a_t \mid s_t) \cdot\frac{1}{\beta} e^{\beta R(t)}d\tau \nonumber \\
&=& \EE{\tau}{\nabla_\theta\log p(\tau\mid\theta)\left(\sum_{t=0}^\infty \nabla_{\theta}\log\pi_{\theta}(a_t \mid s_t) \cdot \frac{1}{\beta} e^{\beta R(t)}\right)^\top} + \EE{\tau}{\sum_{t=0}^\infty \nabla_{\theta}^2\log\pi_{\theta}(a_t \mid s_t) \cdot\frac{1}{\beta} e^{\beta R(t)}} \nonumber \\
&=& \underbrace{\EE{\tau}{\sum_{k=0}^\infty\nabla_\theta\log \pi_\theta(a_k\mid\theta_k)\left(\sum_{t=0}^\infty \nabla_{\theta}\log\pi_{\theta}(a_t \mid s_t) \cdot\frac{1}{\beta} e^{\beta R(t)}\right)^\top}}_{\circled{3}} \nonumber \\
&\quad& + \underbrace{\EE{\tau}{\sum_{t=0}^\infty \nabla_{\theta}^2\log\pi_{\theta}(a_t \mid s_t) \cdot\frac{1}{\beta} e^{\beta R(t)}}}_{\circled{4}}. \label{eq:3+4}
\end{eqnarray}

We also bound the above two terms separately for the risk-sensitive REINFORCE algorithm. 

For the term \circled{3},
\begin{eqnarray}
\norm{\circled{3}} &=& \norm{\EE{\tau}{\sum_{k=0}^\infty\nabla_\theta\log \pi_\theta(a_k\mid\theta_k)\left(\sum_{t=0}^\infty \nabla_{\theta}\log\pi_{\theta}(a_t \mid s_t) \cdot\frac{1}{\beta} e^{\beta R(t)}\right)^\top}} \nonumber \\
&=& \norm{\EE{\tau}{\sum_{k=0}^\infty\nabla_\theta\log \pi_\theta(a_k\mid\theta_k)\left(\sum_{t=0}^\infty \nabla_{\theta}\log\pi_{\theta}(a_t \mid s_t) \cdot\frac{1}{\beta} e^{\beta \sum_{t'=t}^\infty \gamma^{t'} r(s_{t'}, a_{t'})}\right)^\top}} \nonumber \\
&=& \norm{\EE{\tau}{\frac{1}{\beta} e^{\beta \sum_{t=0}^\infty\gamma^t r(s_t, a_t)}\left(\sum_{t'=0}^t\nabla_\theta\log\pi_\theta(a_{t'}\mid \theta_{t'})\right)\left(\sum_{k=0}^t\nabla_\theta\log\pi_\theta(a_k\mid \theta_k)\right)^\top}} \nonumber \\
&\leq& \EE{\tau}{\frac{1}{|\beta|} e^{|\beta| \sum_{t=0}^\infty\gamma^t\left|r(s_t, a_t)\right|}\norm{\sum_{k=0}^t\nabla_\theta\log \pi_\theta(a_{k}\mid\theta_{k})}^2} \nonumber \\
&\overset{\ref{eq: ratio}}{=}& \EE{\tau}{\alpha \sum_{t=0}^\infty\gamma^t|r(s_t, a_t)| \norm{\sum_{k=0}^t\nabla_\theta\log \pi_\theta(a_{k}\mid\theta_{k})}^2} \nonumber \\
&\leq& \alpha \cdot r_{max}\sum_{t=0}^\infty \gamma^t\EE{\tau}{\norm{\sum_{k=0}^t\nabla_\theta\log \pi_\theta(a_{k}\mid\theta_{k})}^2} \nonumber \\
&=& \alpha \cdot r_{max} \sum_{t=0}^\infty \gamma^t \sum_{k=0}^t \EE{\tau}{\norm{\nabla_\theta\log\pi_\theta(a_k\mid\ \theta_k)}^2} \nonumber \\
&\leq& \alpha \cdot r_{max}F_1^2 \sum_{t=0}^\infty \gamma^t(t+1) \nonumber \\
&=& \alpha\cdot\frac{r_{max} F_1^2}{(1-\gamma)^2},
\end{eqnarray}
where in the third line, we use the fact that the future actions do not depend on the past rewards. In the fifth line, we use Assumption \ref{ass: risk-value-ratio}.


For the term \circled{4}, 
\begin{eqnarray}
\norm{\circled{4}} &=& \norm{\EE{\tau}{\sum_{t=0}^\infty \nabla_{\theta}^2\log\pi_{\theta}(a_t \mid s_t) \cdot \beta e^{\beta R(t)}}} \nonumber \\ 
&=& \norm{\EE{\tau}{\sum_{t=0}^\infty \nabla_{\theta}^2\log\pi_{\theta}(a_t \mid s_t) \cdot\frac{1}{\beta} e^{\beta\sum_{t'=t}^\infty \gamma^{t'}r(s_{t'}, a_{t'})}}} \nonumber \\ 
&=& \norm{\EE{\tau} {\frac{1}{\beta} e^{\beta\sum_{t=0}^\infty\gamma^{t}r(s_{t},a_{t})}\left(\sum_{k=0}^t\nabla_{\theta}^2\log\pi_{\theta}(a_k \mid s_k)\right)}} \nonumber \\
&\leq& \EE{\tau} {\frac{1}{|\beta|} e^{|\beta|\sum_{t=0}^\infty\gamma^{t}|r(s_{t},a_{t})|}\left(\sum_{k=0}^t\norm{\nabla_{\theta}^2\log\pi_{\theta}(a_k \mid s_k)}\right)} \nonumber \\
&\overset{\ref{eq: ratio}}{=}& \EE{\tau}{\alpha\cdot\sum_{t=0}^\infty\gamma^t|r(s_t,a_t)|\left(\sum_{k=0}^t\norm{\nabla_{\theta}^2\log\pi_{\theta}(a_k \mid s_k)}\right)} \nonumber \\
&\leq& \alpha \cdot r_{\max}\sum_{t=0}^\infty\gamma^t\left(\sum_{k=0}^t\EE{\tau}{\norm{\nabla_{\theta}^2\log\pi_{\theta}(a_k \mid s_k)}}\right) \nonumber \\ 
&\overset{\ref{eq:E_tau_F2}}{\leq}& \alpha\cdot r_{\max}F_2\sum_{t=0}^\infty\gamma^t(t+1) \nonumber \\
&=& \alpha\cdot\frac{r_{\max} F_2}{(1-\gamma)^2}
\end{eqnarray}

Finally,
\begin{eqnarray}
\norm{\nabla^2J_\beta(\theta)} \leq \alpha\cdot\frac{r_{\max}}{(1-\gamma)^2}(F_1^2+F_2),
\end{eqnarray}
so $L_\beta = \alpha\cdot \frac{r_{\max}}{(1-\gamma)^2}(F_1^2+F_2)$, where $0<\alpha<1$.

\end{document}